\documentclass[preprint,12pt]{elsarticle}

\usepackage{amssymb}
\usepackage{amsmath}

\usepackage{svg}
\usepackage{graphicx}
\usepackage{caption}
\usepackage{subcaption}
\usepackage{makecell}
\usepackage{tabularx, booktabs}
\usepackage{array}
\usepackage{multirow}
\usepackage{float}
\usepackage{longtable}
\usepackage{rotating}
\usepackage{geometry}
\usepackage[decore]{pdflscape}
\usepackage{hyphenat}

\newcommand{\PASS}{\mathrm{PASS}}

\journal{}

\begin{document}

\begin{frontmatter}



\title{Projected Attainable Speed Space: A Driving Efficiency Metric Connecting Instantaneous Evaluation to Travel Time} 


\author[affil1]{Xiaohua Zhao} 
\author[affil1]{Zhaowei Huang}
\author[affil1]{Chen Chen\corref{cor1}}
\author[affil1]{Haiyi Yang}
\cortext[cor1]{Corresponding author. Email: chenchen@bjut.edu.cn}

\affiliation[affil1]{organization={Beijing Key Laboratory of Traffic Engineering, Beijing University of Technology},
            city={Beijing},
            postcode={100124},
            country={P.R.China}}
\begin{abstract}

Inefficient driving behaviors, such as overly conservative yielding, remain a key obstacle to deployment of autonomous vehicles (AVs). Instantaneous driving efficiency metrics are crucial for self-driving decision-making because they affect real-time performance evaluation and control optimization. However, commonly used indicators, including speed, relative speed, and inter-vehicle distance, are limited in capturing traffic context and in ensuring consistency between instantaneous outputs and travel-level outcomes. This study proposes the Projected Attainable Speed Space (PASS) model, a unified framework for driving efficiency assessment across instantaneous and travel-level analyses by integrating kinematic and spatial traffic information. PASS characterizes instantaneous driving efficiency with two coupled elements: potential for speed improvement (available acceleration space) and response to that potential (utilization of available acceleration space). Available acceleration space is referenced to projected attainable speed, derived from an idealized catch-up maneuver using relative speed and spacing to the leading vehicle; utilization is represented by the temporal change in available acceleration space. To ensure cross-scale consistency, time-aggregated PASS is defined as a travel-level efficiency metric. Trajectory data from a driving simulation experiment are used for parameter calibration to maximize agreement between time-aggregated PASS and observed travel times. Across 10 lane-change events, results show strong consistency, with an average coefficient of determination of 0.913, validating PASS for consistent efficiency evaluation across instantaneous and travel-level temporal scales. This study provides a unified, physically grounded framework that supports real-time decision-making and long-term performance analysis in autonomous driving.

\end{abstract}


\begin{highlights}
    \item Proposes PASS model: a unified framework for driving efficiency assessment applicable to both instantaneous and travel-level analyses.
    \item A new concept of available acceleration space is introduced to quantify the potential for speed improvement.
    \item Projects attainable speed based on an idealized catch-up maneuver to provide a physically grounded reference for optimal efficiency.
    \item Consistency between time-aggregated PASS and travel times is validated through a driving simulation experiment.
\end{highlights}


\end{frontmatter}


\section{Introduction}
\label{sec:Intro}
Currently, autonomous vehicles (AVs) are regarded as one of the most promising solutions for enhancing road safety, alleviating traffic congestion, and reducing fuel or energy consumption \cite{ahangar2021survey, ahmed2022technology}. In the foreseeable future, mixed traffic environments, comprising both human-driven vehicles and AVs, are expected to persist \cite{hang2020human}. Within this context, ensuring the safety and driving efficiency of AVs will constitute a central and enduring objective for research and deployment efforts \cite{zhao2024human}.

On-road testing and supporting research have increasingly shown that AV safety performance has substantially improved in recent years \cite{karbasi2022investigating, sourelli2024modelling, wang2024survey}. However, current autonomous driving systems commonly adopt conservative control strategies to maintain a safety margin, which often compromises driving efficiency. For example, \citet{garg2023can} demonstrated that in mixed traffic with a 70\% penetration rate of connected AVs, vehicle conflicts were substantially reduced, but average travel time increased by 2.6\%. Addressing this challenge requires a driving efficiency evaluation framework that can directly support real-time decision-making while remaining meaningful at the travel level. Therefore, a driving efficiency evaluation model that guarantees consistency between in stantaneous correctness and travel-level performance is a prerequisite for driving efficiency optimization in autonomous driving algorithms.

Despite this requirement, the evaluation of driving efficiency in existing studies on AVs has largely relied on travel-level metrics, with total travel time being the most commonly used indicator \cite{zhao2018platoon, guo2019joint, han2020energy, chen2020optimal, pourmehrab2020signalized, li2022analysis, mohebifard2021connected, mohebifard2022trajectory, malikopoulos2018optimal, dai2017effect, sun2023optimal, oh2024exploring}. Although travel time provides a clear and interpretable measure of driving efficiency, it is a travel-level metric that can only be computed upon travel completion and therefore cannot provide feedback during ongoing maneuvers. Consequently, it is incompatible with the requirement for instantaneous driving efficiency evaluation in AVs' real-time decision-making processes.

In response to the limitations of travel-level metrics, several studies have adopted instantaneous speed as a direct metric for instantaneous driving efficiency \cite{zhao2024human, xu2018reinforcement, lv2022safe}. This practice is justified by the ready availability of speed data, its low computational overhead, and its strong empirical association with driving efficiency. However, using speed alone fails to account for the prevailing traffic context. For instance, a vehicle traveling at 30 km/h may be operating near its driving efficiency potential in heavy congestion, yet significantly underperforming in free-flow conditions. Without contextual normalization, such approaches conflate kinematic state with driving efficiency, leading to misleading evaluations across diverse traffic scenarios.

To address the limitation of metrics based solely on instantaneous speed, recent studies have evaluated instantaneous driving efficiency by comparing the ego vehicle's speed with the speeds of surrounding vehicles \cite{hang2020human, hang2021cooperative, yang2023multi, cai2024game}. This relative-speed-based approach allows the driving efficiency metric to distinguish between different traffic contexts, for example by distinguishing between efficient operation in congestion and suboptimal performance in free flow. However, this approach still neglects inter-vehicle distances, which are critical for evaluating the potential for speed improvement. For instance, if the ego vehicle is traveling at a speed similar to that of a leading vehicle but at a large distance, there may be substantial potential to enhance forward progress through safe acceleration, which would not be captured by relative speed alone. Recognizing this deficiency, current research primarily incorporates both relative speed and inter-vehicle distance when evaluating instantaneous driving efficiency. For example, \citet{liu2019novel} developed a benefit metric for discretionary lane change that combines relative speed and inter-vehicle spacing to evaluate instantaneous driving efficiency; higher values of this metric reflect greater potential to improve forward progress under prevailing traffic conditions. \citet{shi2019driving} designed a reward function for hierarchical reinforcement learning-based lane-change decision-making that captures instantaneous driving efficiency by encouraging the ego vehicle to maintain a speed close to that of surrounding traffic while preserving an appropriate inter-vehicle distance that supports sustained forward progress. \citet{lu2024game} constructed a composite payoff within a game-theoretic lane-change decision framework that quantifies instantaneous driving efficiency by integrating relative speed and inter-vehicle distance to reflect the potential for increasing velocity and available spatial freedom for smooth progression. \citet{deng2022lane} proposed a Bayesian game-based lane-change model for dense highway traffic in which instantaneous driving efficiency is evaluated through the speed advantage over a leading vehicle together with the available acceleration space in the current lane, the latter being quantified by the distance to the following vehicle to avoid unnecessary maneuvers when forward progress can be improved without changing lanes. Nevertheless, these approaches are mostly application-specific and do not provide a unified framework that evaluates driving efficiency under instantaneous traffic states while maintaining consistency with travel-level outcomes across maneuvers and traffic contexts.

To address this gap, this study proposes the Projected Attainable Speed Space (PASS) model, which provides a unified framework for instantaneous driving efficiency evaluation that is applicable across diverse maneuvers and traffic contexts. The conceptual basis of PASS is that driving efficiency should be judged under the current traffic-flow context, and it is framed by two coupled questions: whether the ego vehicle still has potential to improve speed, and whether the vehicle responds in a way that utilizes that potential. By jointly considering these two aspects, the proposed framework supports instantaneous driving efficiency evaluation while remaining extendable to travel-level driving efficiency evaluation. Through parameter calibration using vehicle trajectory data collected from a driving simulation experiment, the time-aggregated PASS is shown to exhibit strong consistency with observed travel times, validating the model's ability to provide consistent driving efficiency evaluations across instantaneous and travel-level temporal scales. The main contributions of this work are summarized as follows:

\begin{itemize}
  \item A conceptual framework for instantaneous driving efficiency evaluation is proposed, which characterizes driving efficiency through the potential for speed improvement and its utilization, providing a comprehensive and dynamic understanding of instantaneous driving efficiency.  
  \item An instantaneous driving efficiency evaluation model is developed based on the proposed framework. The PASS model integrates both kinematic and spatial information from the surrounding traffic environment to provide a unified and physically grounded evaluation of instantaneous driving efficiency.
  \item The time-aggregated PASS exhibits strong consistency with travel-level driving efficiency metrics, confirming that the proposed model provides consistent driving efficiency evaluations across instantaneous and travel-level temporal scales.
\end{itemize}

\section{The Projected Attainable Speed Space (PASS) Model}
\label{sec:PASS}

\subsection{Conceptual Framework}
\label{ssec:conceptual_framework}
The highest-level premise of the PASS concept is that driving efficiency should be judged by a vehicle's performance under its current traffic-flow context, rather than by speed alone. In other words, driving efficiency is inherently contextual: the same speed may indicate efficient operation in one traffic state but inefficient operation in another. Based on this premise, the conceptual evaluation is decomposed into two coupled questions: whether the ego vehicle still has potential to improve speed, and whether the vehicle responds in a way that utilizes that potential. 

By jointly considering potential for speed improvement, which is quantified as the available acceleration space, and the response of the vehicle to that potential, which is quantified as the utilization of the available acceleration space, PASS establishes a unified conceptual basis for instantaneous driving efficiency evaluation across different maneuvers and traffic contexts. Starting from this instantaneous-evaluation objective, the framework further requires that aggregated instantaneous evaluations remain consistent with travel-level driving efficiency evaluation, which motivates the subsequent model construction, metric formulation, and calibration strategy presented in the following subsections.

\begin{figure}[htb]
    \centering
    \includegraphics[width=\linewidth]{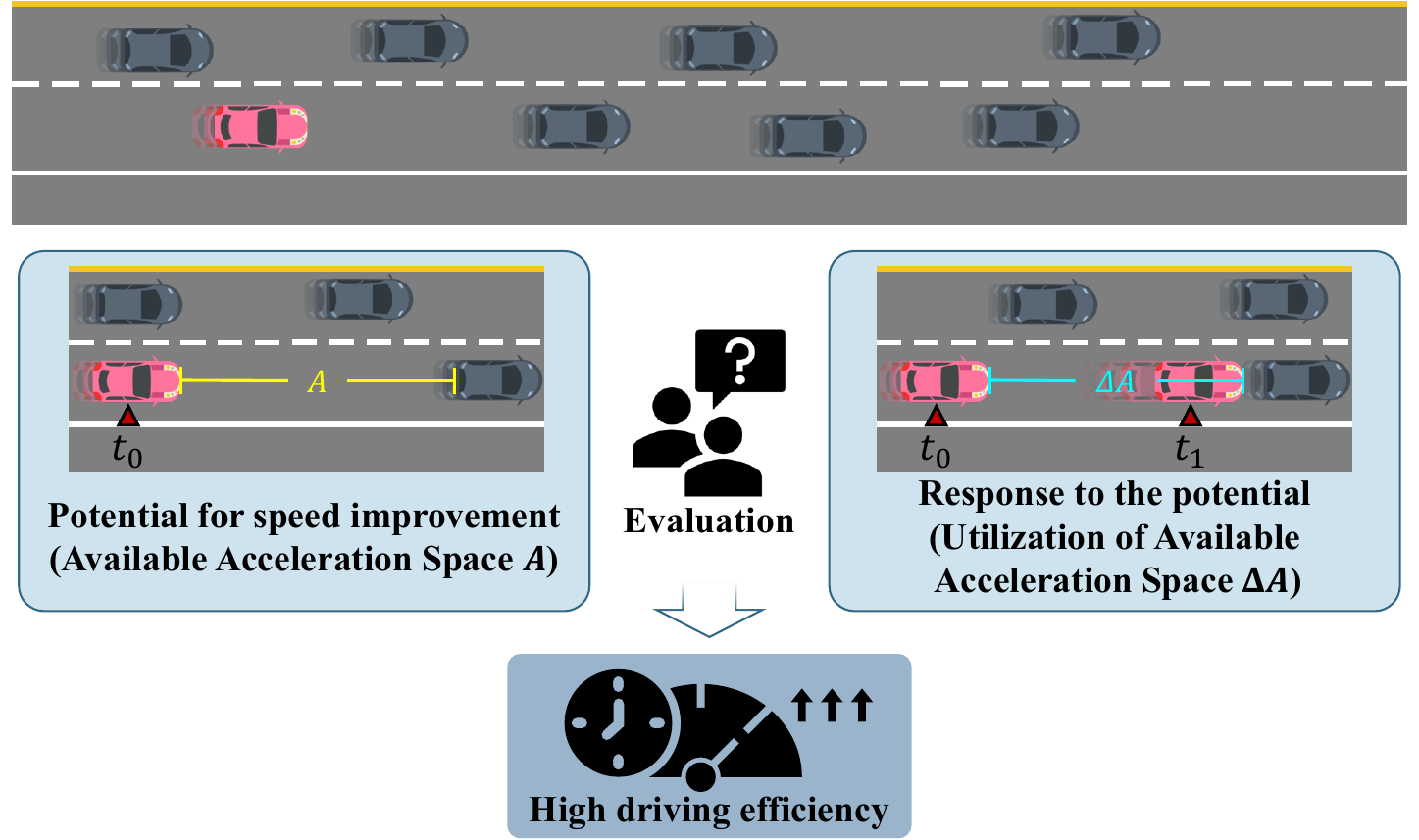}
    \caption{Conceptual framework of the PASS model}
    \label{fig:concept}
\end{figure}

\subsection{Available Acceleration Space}
\label{ssec:available_acceleration_space}
To quantify the potential for speed improvement under the current traffic context, the PASS model introduces the concept of available acceleration space. This metric represents the potential for speed improvement by quantifying the deviation of the ego vehicle's current speed from an optimal reference speed that reflects the best attainable speed under prevailing traffic conditions. Specifically, the available acceleration space at time $t$ is defined as the difference between the optimal reference speed $v_{\text{opt},t}$ and the ego vehicle's instantaneous speed $v_{0,t}$:
\begin{equation}
    A_t = v_{\text{opt},t} - v_{0,t}.
    \label{eq:A_t_concept}
\end{equation}

The interpretation of the available acceleration space $A_t$ is as follows:
\begin{itemize}
    \item If $A_t > 0$, the ego vehicle's current speed is below the optimal reference speed, indicating the presence of untapped potential for speed improvement; greater values of $A_t$ correspond to lower instantaneous driving efficiency; 
    \item If $A_t \leq 0$, the ego vehicle's current speed meets or exceeds the optimal reference speed, indicating that the available acceleration space has been effectively utilized and the vehicle operates at a high level of instantaneous driving efficiency.
\end{itemize}

\subsubsection{Projected Attainable Speed}
\label{sssec:projected_attainable_speed}
To establish a physically grounded optimal reference speed, the PASS model introduces the concept of projected attainable speed. This metric represents the theoretical maximum average speed that an ego vehicle can attain over a short prediction horizon under current traffic conditions. The projected attainable speed is derived from an idealized catch-up maneuver, which characterizes the process by which a vehicle maximizes its driving efficiency through close following of the leading vehicle. This maneuver assumes surrounding vehicles maintain their current kinematic states (i.e., constant speed and lane position), enabling real-time evaluation while jointly accounting for relative kinematics and inter-vehicle spacing.

\begin{figure}[!h]
    \centering
    \includegraphics[width=\linewidth]{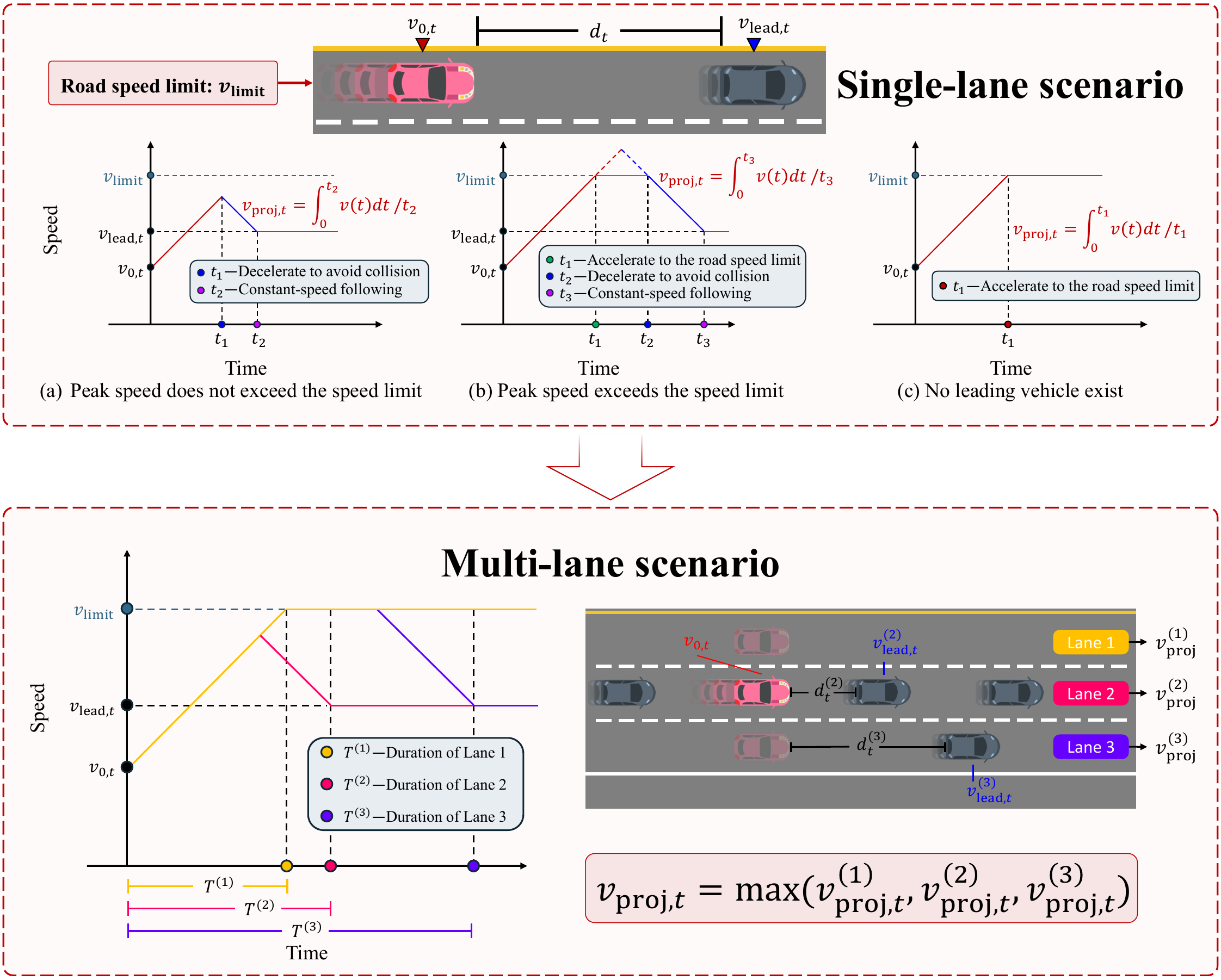}
    \caption{Illustration of the projected attainable speed}
    \label{fig:v_proj}
\end{figure}

Figure~\ref{fig:v_proj} summarizes the speed profiles of the idealized catch-up maneuver in the single-lane context, and the calculation of the projected attainable speed in the multi-lane context. The following sections provide detailed descriptions of the projected attainable speed calculation in both single-lane and multi-lane contexts.

\paragraph{\textbf{Single-lane Scenario}}
\label{ssssec:single_lane}
In a single-lane context, the projected attainable speed is computed based on an idealized catch-up maneuver that consists of three sequential phases: (1) acceleration to close the gap, (2) deceleration to avoid collision, and (3) cruising at the same speed as the leading vehicle with minimal inter-vehicle spacing. In the limiting case of maximal driving efficiency, the final spacing is permitted to approach zero. 

The idealized catch-up maneuver is defined under the following assumptions:
\begin{itemize}
    \item The leading vehicle maintains constant speed and lane position throughout the maneuver.
    \item The ego vehicle uses constant acceleration during acceleration phases and constant deceleration during deceleration phases.
\end{itemize}

Given the current traffic state, which includes the ego vehicle's speed, the vehicle spacing to the leading vehicle, the leading vehicle's speed, and the road speed limit, the projected attainable speed is computed as the average speed over the duration of the idealized catch-up maneuver. Two cases are distinguished based on whether the peak speed reached during the maneuver exceeds the speed limit.

Case 1: peak speed does not exceed the speed limit (Fig.~\ref{fig:v_proj}(a)). If the vehicle does not reach the speed limit during the maneuver, the projected attainable speed is given by
\begin{equation}
    v_{\text{proj},t} = \frac{a_1a_2d_t}{a_2(v_{\text{lead},t}-v_{0,t})+(a_2-a_1)\sqrt{\frac{a_2}{a_2-a_1}[2a_1d_t + (v_{\text{lead},t}-v_{0,t})^2]}} + v_{\text{lead},t}
    \label{eq:vproj_case1}
\end{equation}
where $ v_{\text{proj},t} $ denotes the projected attainable speed at time $t$, $ d_t $ is the ego vehicle's spacing to the leading vehicle at time $t$, $ v_{0,t} $ is the ego vehicle's current speed at time $t$, $ v_{\text{lead},t} $ is the leading vehicle's speed at time $t$, $ a_1 $ is the constant acceleration during the acceleration phase, and $ a_2 $ is the constant deceleration during the deceleration phase.

Case 2: peak speed exceeds the speed limit (Fig.~\ref{fig:v_proj}(b)). If the acceleration phase would cause the vehicle to exceed the speed limit, a cruising phase at the speed limit is inserted between acceleration and deceleration. The projected attainable speed is then
\begin{equation}
    v_{\text{proj},t} = \frac{2a_1a_2d_t(v_{\text{limit}}-v_{\text{lead},t})}{2a_1a_2d_t+a_2(v_{\text{limit}}-v_{0,t})^2-a_1(v_{\text{limit}}-v_{\text{lead},t})^2} + v_{\text{lead},t}
    \label{eq:vproj_case2}
\end{equation}
where $ v_{\text{limit}} $ denotes the road speed limit.

By default, the acceleration and deceleration parameters are set to  $ a_1 = 1.5~\mathrm{m/s^2} $  and  $ a_2 = -1.5~\mathrm{m/s^2} $ , representing typical passenger vehicle capabilities. However, these values can be adjusted based on vehicle dynamics or safety policies.

Special cases are handled as follows:
\begin{itemize}
    \item \textbf{No leading vehicle exists:} As shown in Fig.~\ref{fig:v_proj}(c), when no vehicle is present ahead, the catch-up maneuver reduces to accelerating to the speed limit and cruising thereafter. The projected attainable speed approaches the speed limit asymptotically over the prediction horizon.
    
    \item \textbf{Static obstacle exists:} A stationary obstacle is treated as a leading vehicle with  $ v_{\text{lead},t} = 0 $ . If both a leading vehicle and a static obstacle are present ahead in the same lane $ i $, calculate the projected attainable speeds of the leading vehicle ($v_{\text{proj},t}^{(i,\text{lead})}$) and the static obstacle ($v_{\text{proj},t}^{(i,\text{static})}$) separately, and the more restrictive value is selected: 
    \begin{equation}
        v_{\text{proj},t} = \min\left( v_{\text{proj},t}^{(i,\text{lead})},\; v_{\text{proj},t}^{(i,\text{static})} \right). 
        \label{eq:vproj_obstacle}
    \end{equation}

    \item \textbf{Ego vehicle's speed is too high:} If the ego vehicle's current speed is too high relative to the leading vehicle, such that it would need to decelerate immediately to avoid collision, the acceleration phase of the catch-up maneuver is omitted. In this case, the projected attainable speed is computed based solely on the deceleration phase.
    
    \item \textbf{Adjustment to avoid collision:} If the nominal deceleration  $ a_2 = -1.5~\mathrm{m/s^2} $  cannot ensure collision avoidance given the current speed and spacing, i.e., if the required stopping distance exceeds the available spacing, then the deceleration is adjusted to ensure kinematic feasibility. Specifically, if
    \begin{equation}
        \frac{(v_{0,t} - v_{\text{lead},t})^2}{2 |a_2|} > d_t,
        \label{eq:a2_condition}
    \end{equation}
    then the deceleration is increased to the minimum value that ensures kinematic feasibility:
    \begin{equation}
        a_2^{\text{adj}} = -\frac{(v_{0,t} - v_{\text{lead},t})^2}{2 d_t}.   
        \label{eq:a2_adj}     
    \end{equation}
    This adjusted deceleration $ a_2^{\text{adj}} $ is used in place of the nominal  $ a_2 $  in both Eq.~\eqref{eq:vproj_case1} and Eq.~\eqref{eq:vproj_case2}.
\end{itemize}

\paragraph{\textbf{Multi-lane Scenario}}
\label{ssssec:multi_lane}
In a multi-lane context, the ego vehicle can enhance its driving efficiency by selecting the optimal target lane prior to executing the catch-up maneuver. The projected attainable speed is defined as the maximum average speed attainable across all candidate lanes, where the averaging is performed over a common time horizon equal to the longest duration among all individual catch-up maneuvers.

The projected attainable speed in the multi-lane context is then given by
\begin{equation}
    v_{\text{proj},t} = \max_{i \in \mathcal{I}} \, v_{\text{proj},t}^{(i)}
    \label{eq:vproj_multi}
\end{equation}
\begin{equation}
    v_{\text{proj},t}^{(i)} = \frac{D_{\text{total}}^{(i)}}{T_{\max}}
    \label{eq:vproj_lanek}
\end{equation}
\begin{equation}
    D_{\text{total}}^{(i)} = D^{(i)} + v_{\text{lead},t}^{(i)} \bigl( T_{\max} - T^{(i)} \bigr)
    \label{eq:D_total}
\end{equation}
\begin{equation}
    T_{\max} = \max_{i \in \mathcal{I}} T^{(i)}
    \label{eq:T_max}
\end{equation} 
where $ \mathcal{I} $  is the set of candidate lanes, $ T^{(i)} $  is the duration of the idealized catch-up maneuver in lane  $ i $, $ D^{(i)} $  is the distance traveled during the catch-up maneuver in lane  $ i $, and  $ v_{\text{lead},t}^{(i)} $  is the leading vehicle's speed in lane  $ i $ at time $ t $, $ D_{\text{total}}^{(i)} $  is the total distance traveled in lane  $ i $  over the unified time horizon  $ T_{\max} $, and  $ v_{\text{proj},t}^{(i)} $  is the average speed in lane  $ i $  over the time horizon  $ T_{\max} $.

\subsection{Utilization of Available Acceleration Space}
\label{ssec:utilization}
While the available acceleration space $A_t$ provides an instantaneous measure of the potential for speed improvement, it still needs to capture the response of the vehicle to this potential over time. To evaluate the response, the utilization of the available acceleration space is defined based on its discrete temporal change:

\begin{equation}
    \Delta A_t = A_t - A_{t-1}.
    \label{eq:delta_A}
\end{equation}

The sign of $\Delta A_t$ indicates the direction of change in instantaneous driving efficiency:

\begin{itemize}
    \item $\Delta A_t < 0$: The available acceleration space is decreasing, indicating that the vehicle is actively utilizing the existing potential for speed improvement, thus moving towards the optimal reference.
    \item $\Delta A_t > 0$: The available acceleration space is increasing, suggesting that the current driving state is drifting further from the optimal reference, thus moving away from optimal reference speed.
    \item $\Delta A_t = 0$: The available acceleration space remains unchanged, indicating that the vehicle is maintaining the current level of driving efficiency without actively improving or deteriorating it.
\end{itemize}

To allow for different responsiveness to temporal changes in the high-efficiency regime ($A_t \leq 0$) and the low-efficiency regime ($A_t > 0$), two distinct scaling coefficients are introduced:

\begin{equation}
    {\Delta A_t}^{\prime} =
    \begin{cases}
        k_1 \Delta A_t, & \text{if } A_t \leq 0, \\
        k_2 \Delta A_t, & \text{if } A_t > 0,
    \end{cases}
    \label{eq:f_deltaA}
\end{equation} 
where $k_1 < 0 < k_2$ are the response coefficients to be calibrated.

\subsection{PASS Metric}
\label{ssec:PASS_metric}
\subsubsection{Instantaneous Metric}
\label{sssec:instantaneous_metric}
By integrating the available acceleration space $A_t$ and its utilization ${\Delta A_t}^{\prime}$, the instantaneous metric PASS can be defined as:
\begin{equation}
    \PASS_t = A_t \left[ \tanh\!\big( {\Delta A_t}^{\prime} \big) + 1 \right].
    \label{eq:PASS_t}
\end{equation}
The hyperbolic tangent function ensures smooth and bounded scaling within the interval [0, 2].

Larger values of $\PASS_t$ consistently indicate greater deviation from the optimal reference speed, and thus lower instantaneous driving efficiency, preserving the monotonic relationship inherited from $A_t$.

\subsubsection{Travel-level Metric}
\label{sssec:travel_level_metric}
To verify the effectiveness of the instantaneous PASS metric, it is necessary to ensure that it provides consistent evaluations across different temporal scales. Specifically, if the instantaneous PASS metric is effective, then its time-aggregated value over a travel should exhibit a strong positive association with observed travel times. To this end, the time-aggregated PASS is computed as the average of instantaneous PASS values over the duration of a travel:

\begin{equation}
    \overline{\PASS} = \frac{1}{N} \sum_{t=1}^{N} \PASS_{t}.
    \label{eq:avg_PASS}
\end{equation}
where $N$ is the total number of time steps in the travel and $t$ is the time-step index.

By calibrating the parameters $k_1$ and $k_2$ to maximize the consistency between $\overline{\PASS}$ and observed travel times, the model's ability to provide consistent driving efficiency evaluations across instantaneous and travel-level temporal scales can be validated.

\section{Experiment}
\label{sec:experiment}
\subsection{Scenario Design}
\label{ssec:scenario_design}
This study aims to develop an instantaneous driving efficiency evaluation model for AVs and calibrate its parameters using trajectory-level driving behavior data. A driving simulation experiment is conducted to collect vehicle trajectory data during mandatory lane change (MLC) scenarios. Such scenarios represent strong strategic interactions among vehicles, where drivers must balance driving efficiency, safety, and maneuver feasibility under tight spatiotemporal constraints. As a result, driving efficiency exhibits substantial variation under these conditions, providing suitable data for model calibration.

\begin{figure}[htb]
    \centering
    \includegraphics[width=\linewidth]{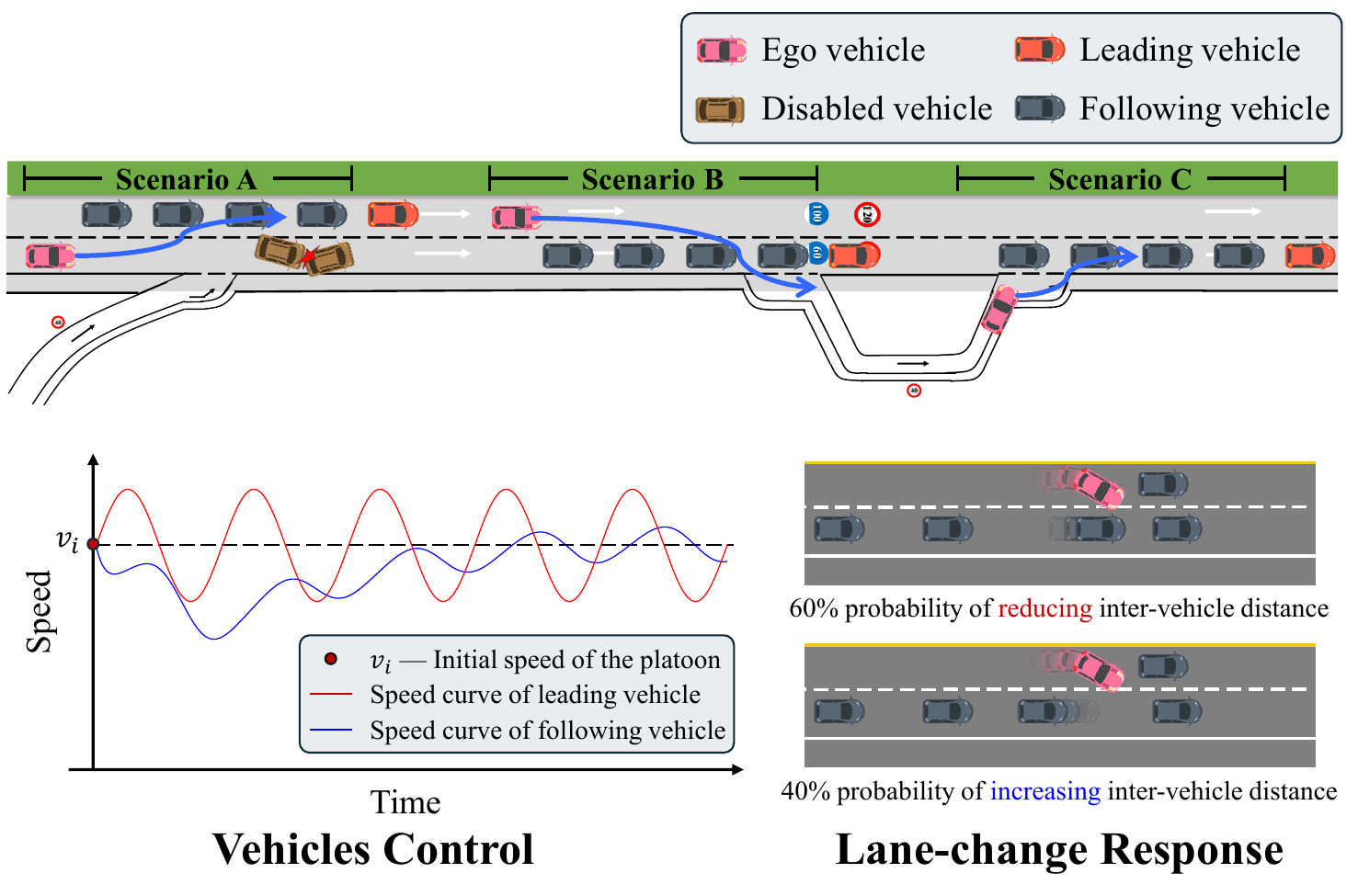}
    \caption{Illustration of the three representative MLC scenarios used in the driving simulation experiment: (Scenario A) Incident avoidance, (Scenario B) Service area entry via off-ramp, and (Scenario C) Highway entry from on-ramp.}
    \label{fig:scenarios}
\end{figure}

Three representative MLC scenarios are selected to capture trajectory data in which instantaneous driving efficiency exhibits substantial variation. As illustrated in Fig.~\ref{fig:scenarios}, the scenarios are:

\begin{enumerate} 
\item \textbf{Incident Avoidance (Scenario A):} The ego vehicle encounters a disabled vehicle blocking the current lane ahead and must execute a MLC to avoid collision. 
\item \textbf{Service Area Entry (Scenario B):} The ego vehicle must enter a service area via an off-ramp and performs a MLC before reaching the ramp entrance to avoid missing the exit. 
\item \textbf{Highway Entry (Scenario C):} The ego vehicle merges onto a highway from an on-ramp and performs a MLC into the mainline traffic within a limited distance to avoid stopping at the end of the acceleration lane. 
\end{enumerate}

In all three scenarios, the ego vehicle is required to complete a MLC under external spatial and temporal constraints. These constraints lead to significant changes in instantaneous driving efficiency, yielding trajectory data that support the calibration of the PASS across diverse traffic contexts. 

\subsection{Platoon Vehicle Control Mechanism}
\label{ssec:vehicle_control}
The driving simulation experiment is conducted in the SCANeR driving simulation platform, which provides high-fidelity visual, auditory, and vehicle dynamics feedback. However, the default traffic model in SCANeR assumes cooperative behavior: surrounding vehicles automatically yield to merging vehicles once a predefined time-to-collision threshold is breached. This assumption fails to capture the competitive or non-yielding behaviors commonly observed in real-world dense traffic, where drivers may deliberately reduce headway to block unwanted merges.

To address this limitation, a custom vehicle control framework is implemented via SCANeR's Python API. The framework controls the behavior of vehicles in the target lane using three key mechanisms:

\begin{enumerate} 
\item \textbf{Lead Vehicle Control:} Instead of maintaining constant speed, the target lane's lead vehicle's velocity is modulated by a low-amplitude sinusoidal signal within a prescribed range. This introduces realistic speed fluctuations that propagate through the platoon, mimicking natural traffic oscillations. The base speed of the lead vehicle is set within the range of 20-40 km/h, depending on the scenario.
\item \textbf{Following Vehicle Control:} The Intelligent Driver Model (IDM) is employed to govern the longitudinal behavior of following vehicles in the target lane. By adjusting acceleration and deceleration based on the relative speed and spacing to the lead vehicle, IDM allows for realistic car-following dynamics. The headway of each vehicle is maintained between 1.0 and 2.0 s.
\item \textbf{Probabilistic Lane-change Response:} Upon detecting an ego vehicle attempting to merge, the following vehicle in the target lane independently decides whether to actively reduce inter-vehicle distance with a certain probability. If triggered, the vehicle temporarily decreases its minimum time headway, thereby increasing resistance to the merge attempt. 
\end{enumerate}

\subsection{Participants and Data Filtering}
\label{ssec:participants}
\begin{figure}[htb]
    \centering
    \includegraphics[width=\linewidth]{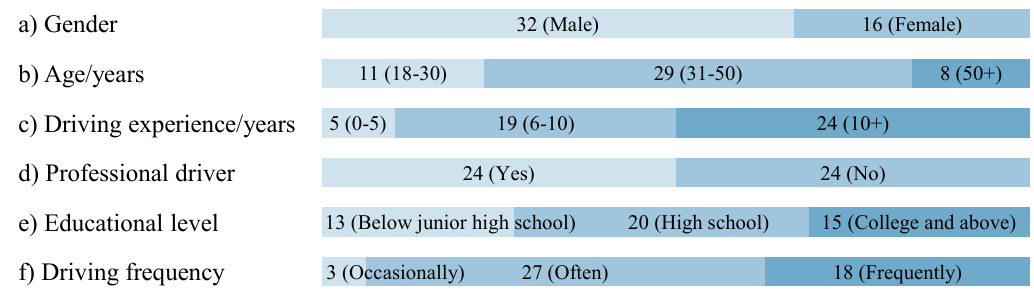}
    \caption{Participant demographics}
    \label{fig:demographics}
\end{figure}

A total of 48 participants (32 male, 16 female) were recruited from universities and social organizations. The gender distribution aligns approximately with national statistics for licensed drivers in China. All participants held valid driver's licenses, had normal vision and hearing, and passed a pre-test confirming their ability to operate the driving simulator. Demographic characteristics are summarized in Fig.~\ref{fig:demographics}.

Trajectory data were recorded at a sampling frequency of 20 Hz throughout each trial. The recorded variables include timestamp, lane ID, global coordinates $(x, y)$, speed, and acceleration.

Five participants were excluded from analysis because they were unable to complete the experimental process due to simulator sickness or protocol deviation. To ensure the validity of the results, the required sample size was calculated based on expected variance, a target confidence level, and an acceptable margin of error \cite{bian2021influence}. The final sample size ($n = 43$) meets the requirements for reliable estimation.

\subsection{Parameter Calibration}
\label{ssec:parameter_calibration}
For each of the 10 lane-change events, time-aggregated PASS and observed travel time were computed for all participating vehicles. Within each event, both variables were ranked, and the Spearman rank correlation coefficient $r_e$ between the two ranked vectors was calculated. The corresponding rank-based coefficient of determination, $R_e^2 = r_e^2$, was used to evaluate how well PASS preserves the same driving-efficiency ordering as observed travel time under identical traffic conditions.

The response coefficients $k_1$ and $k_2$ in the PASS model were calibrated using the composite loss function defined as follows:

\begin{equation}
    \mathcal{L}(k_1, k_2) = \sum_{e \in \mathcal{E}} \left( 1 - r_e^2 
    + \mathbb{I}(r_e < 0) \cdot 10 \cdot |r_e| 
    + \mathbb{I}(r_e^2 < 0.8) \cdot 10 \cdot (0.8 - r_e^2)^2 \right),
    \label{eq:calibration_loss}
\end{equation}

where $\mathcal{E}$ is the set of all lane-change events in the driving simulator experiment, and $r_e$ is the Spearman rank correlation coefficient between time-aggregated PASS and travel time for event $e$.

This formulation explicitly enforces three requirements: a positive association between time-aggregated PASS and observed travel time ($r_e > 0$), a strong within-event relationship ($R_e^2 \to 1$), and robust alignment across all events. Parameter optimization was conducted via grid search over a predefined parameter space, with $k_1 \in [-1,0]$, $k_2 \in [0,1]$, and a step size of 0.01. The optimal parameter pair was selected by minimizing the loss function, corresponding to the best overall consistency between time-aggregated PASS and observed travel times across all events. The calibration procedure was implemented in Python using SciPy for optimization and statistical analysis.

\section{Results}
\label{sec:results}
The response coefficients in the PASS model were calibrated using the composite loss function defined in Eq.~\eqref{eq:calibration_loss}. The optimization yielded the final parameter values:
\[
k_1 = -0.417, \quad k_2 = 0.700.
\]

To evaluate calibration performance, the rank correlation between time-aggregated PASS ($\overline{\PASS}$) and travel time was assessed across the 10 lane-change events used for calibration. For each event, vehicles were ranked separately by $\overline{\PASS}$ and by travel time, and the coefficient of determination $R^2$ was computed. 

\begin{figure}[htb]
    \centering
    \includegraphics[width=\linewidth]{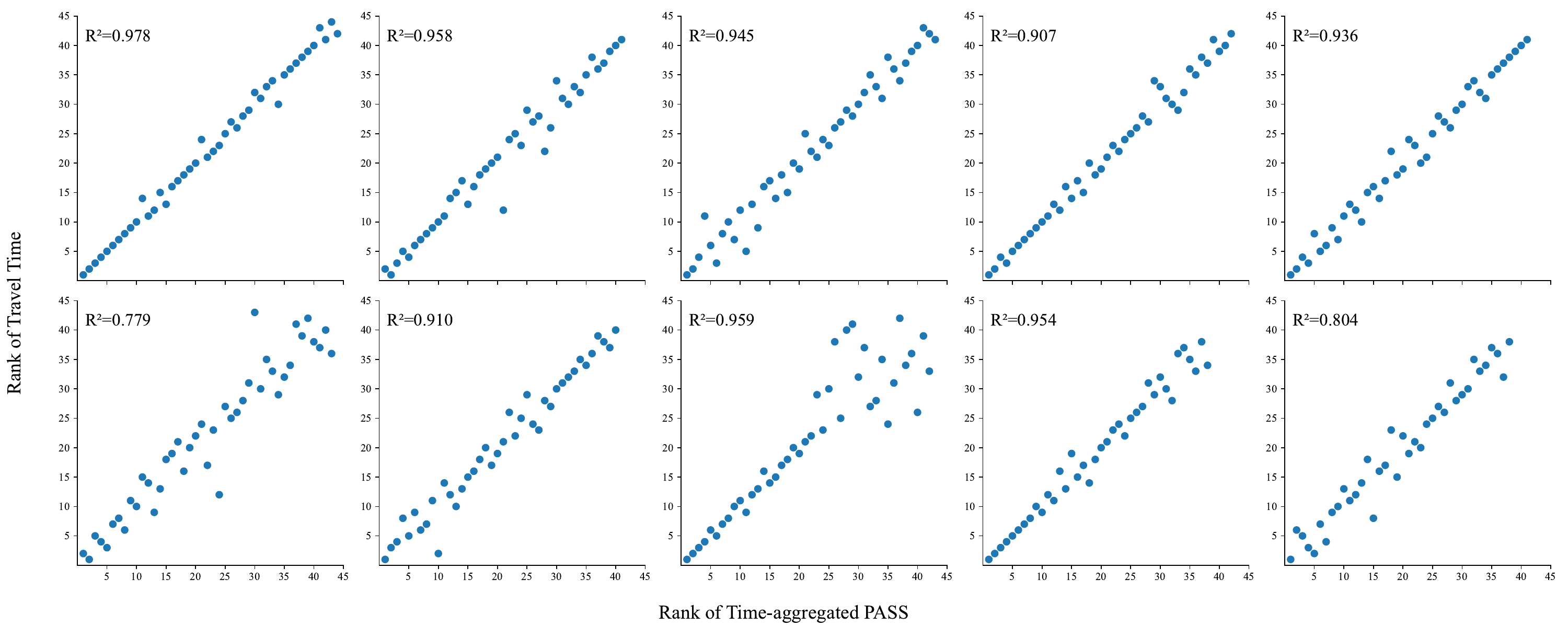}
\caption{Scatter plots of ranked $\overline{\PASS}$ (horizontal axis) versus ranked travel time (vertical axis) across all 10 calibration events. Each point represents one participant vehicle. The high $R^2$ values indicate strong alignment between $\overline{\PASS}$ and travel time in evaluating travel-level driving efficiency.}
\label{fig:calibration_scatter}
\end{figure}

Fig.~\ref{fig:calibration_scatter} shows the scatter plots of ranked $\overline{\PASS}$ versus ranked travel time for all 10 events. Most events exhibit a strong positive association, with 8 out of 10 events achieving coefficients of determination above 0.90 ($R^2 > 0.90$), and the average $R^2$ across all events is 0.913, indicating strong agreement between $\overline{\PASS}$ and travel time. This high level of consistency confirms that $\overline{\PASS}$ effectively evaluates driving efficiency at the travel level, supporting its validity as a meaningful metric.

\section{Discussion}
\label{sec:discussion}
\subsection{Application in Self-driving algorithm design}
\label{ssec:discussion_contribution}
Existing instantaneous metrics used in the reward functions of reinforcement learning-based decision-making algorithms of AVs are solely focused on evaluating the current kinematic state of the vehicle, without guaranteeing that their temporal aggregation will align with travel-level efficiency evaluation. Specifically, while these metrics may provide meaningful feedback for real-time control decisions, they do not ensure that the same trajectory that achieves the best travel time will also obtain the best instantaneous efficiency rank at each moment. As a result, these metrics may lead to suboptimal decision-making when aggregated over time, as they can incentivize actions that improve instantaneous efficiency but do not contribute to overall travel efficiency. 

Compared with commonly used instantaneous metrics, PASS explicitly embeds a consistency requirement between instantaneous evaluation and travel-level outcomes. The PASS model is designed to ensure that its time-aggregated value ($\overline{\PASS}$) exhibits a strong positive association with observed travel times, thereby providing a consistent evaluation framework across temporal scales. This consistency is crucial for developing AV decision-making algorithms that optimize for long-term efficiency while still providing meaningful instantaneous feedback. By calibrating the response coefficients to maximize the alignment between $\overline{\PASS}$ and travel time, the model ensures that the same trajectory that achieves the best travel time also obtains the best $\overline{\PASS}$ rank, thus reinforcing the connection between instantaneous efficiency evaluation and travel-level outcomes. This contribution is significant as it addresses a critical gap in the design of instantaneous driving efficiency metrics, providing a more robust and reliable framework for evaluating and optimizing AV performance in real-world traffic conditions.

\subsection{Comparison with an Existing Metric}
\label{ssec:discussion_comparison}
To validate the effectiveness of the PASS model in evaluating instantaneous driving efficiency and its consistency with travel-level efficiency, it is important to compare its performance against an existing instantaneous metric. Therefore, a comparative analysis was conducted between PASS and one baseline metric (based on the study by \citet{liu2019novel}) that considers the ego vehicle's speed relative to the front vehicle's speed and the inter-vehicle spacing, which is commonly used in the decision-making algorithms of AVs. Similar to PASS, the magnitude (0 to 1) of the baseline metric is negatively associated with instantaneous driving efficiency.


To illustrate the comparative performance of PASS and the baseline metric, a representative incident avoidance scenario event (event 1) is examined. In this event, a downstream lane closure is caused by two stationary and visibly damaged vehicles blocking Lane 1, requiring all approaching vehicles to perform a MLC into Lane 0 before reaching the incident site. The target lane (Lane 0) carries a congested traffic stream, where vehicle speeds exhibit natural fluctuations.

\begin{figure}[htb]
    \centering
    \includegraphics[width=\linewidth]{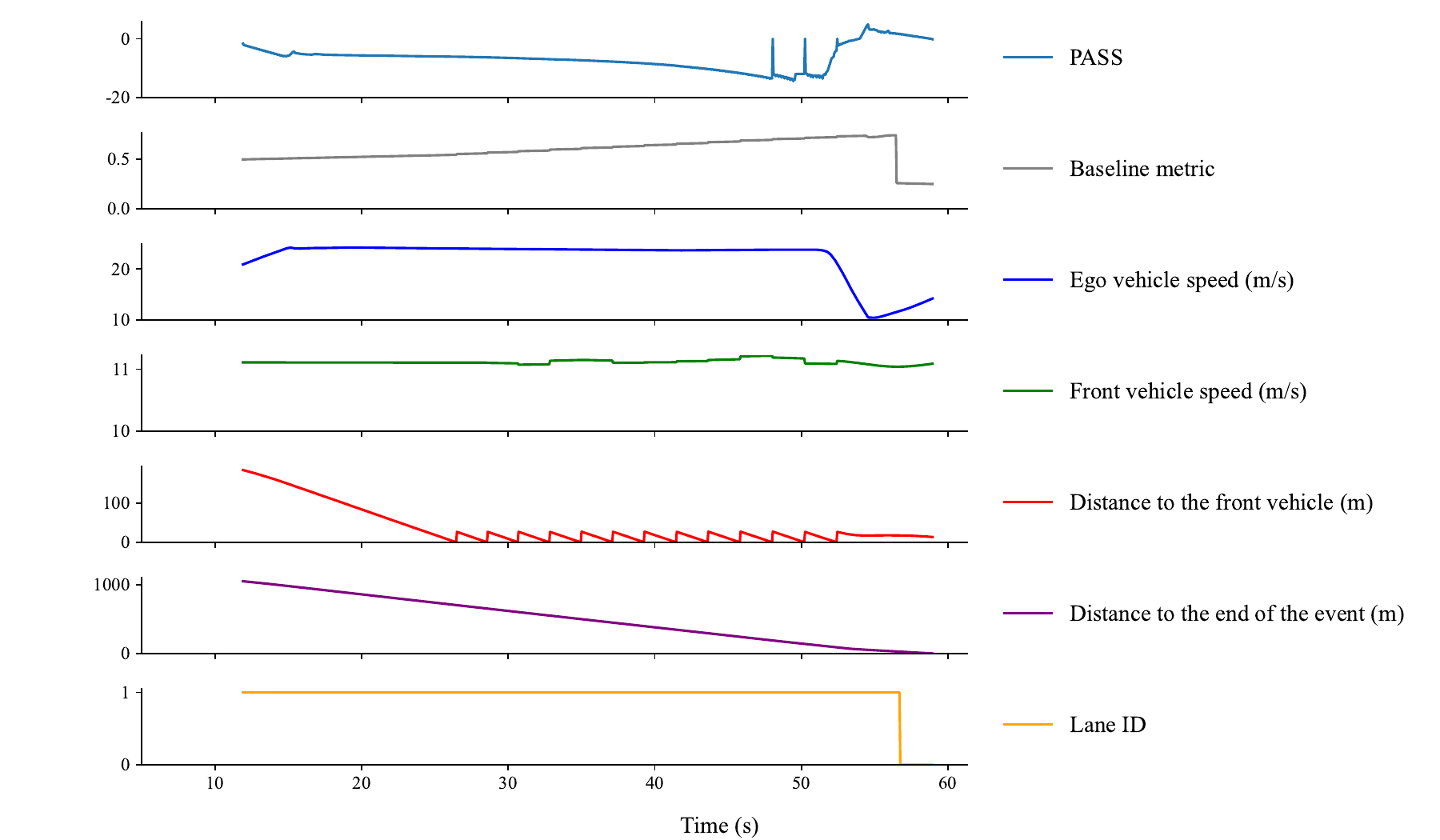}
\caption{Time series of driving dynamics for the top-ranked vehicle in Event 1}
\label{fig:event1_vehicle1}
\end{figure}

Event 1 exhibits a high coefficient of determination ($R^2 = 0.978$) between the ranks of $\overline{\PASS}$ and travel time. One participant vehicle achieved rank 1 in both metrics, corresponding to the shortest travel time and the lowest $\overline{\PASS}$, indicating the most efficient maneuver observed under these constraints. Fig.~\ref{fig:event1_vehicle1} presents the time series of key variables for this vehicle, including instantaneous PASS ($\PASS_t$), baseline metric, ego vehicle speed, front vehicle speed, distance to the front vehicle, distance to the end of event 1, and lane ID.

Throughout most of the event duration, the PASS remained consistently low (below $-10$), reflecting sustained high driving efficiency in the uncongested lane. As the vehicle neared the incident site, the projected attainable speed $v_{\text{proj},t}$ in Lane 1 decreased due to proximity to the static obstruction. At approximately $48~\mathrm{s}$, the PASS model switched its reference to the target lane (Lane 0), selecting $v_{\text{proj},t}^{(0)}$ as the reference projected attainable speed because it became larger than the $v_{\text{proj},t}^{(1)}$. This transition caused an abrupt change in the available acceleration space, resulting in a sharp, transient increase in the instantaneous PASS. This abrupt change reflects the model's immediate response to a new traffic context, specifically a shift in the evaluation reference from a static obstacle to the platoon. In contrast, the baseline metric remained comparatively high (above $0.5$) for a significant portion of the event, indicating lower instantaneous efficiency according to the baseline metric. As the vehicle neared the incident site, the baseline metric gradually increased until the MLC was completed and then significantly decreases. This significant difference is due to the difference in the treatment of the static obstacle in the two models. The baseline metric can not capture the potential for speed improvement in the current lane due to the static obstacle, it only considers that the static obstacle  will reduce the final speed of the ego vehicle in the current lane to 0, but does not consider that the ego vehicle can still achieve a high speed by performing a MLC to the target lane later. In contrast, the PASS model captures the potential for speed improvement in the current lane even in the presence of the static obstacle, because the projected attainable speed in the current lane is still high at the beginning of the event when the ego vehicle is far from the incident site. This allows the PASS model to provide a more accurate and dynamic evaluation of instantaneous driving efficiency.

From $50~\mathrm{s}$ onward, the vehicle began decelerating to merge into the congested stream in Lane 0, causing a gradual rise in the PASS as speed decreased and headway compressed. However, the baseline metric continued to increase slowly, failing to capture the decrease in instantaneous driving efficiency during the deceleration and merging process. The lane change was completed around $56~\mathrm{s}$, after which the ego vehicle accelerated smoothly within the target lane. Correspondingly, PASS and the baseline metric declined again, but the decline in PASS was more pronounced, reflecting the significant improvement in driving efficiency after merging into the target lane. Notably, the speed profile of the leading vehicle in Lane 0 follows a sinusoidal pattern, confirming that the ego vehicle successfully inserted itself ahead of the entire platoon, becoming the immediate follower of the platoon leader. This strategic choice to delay lane change until the last moment minimizes time spent in slow platoon, explaining why this trajectory achieved rank first in both travel time and $\overline{\PASS}$.

\begin{figure}[htb]
    \centering
    \includegraphics[width=\linewidth]{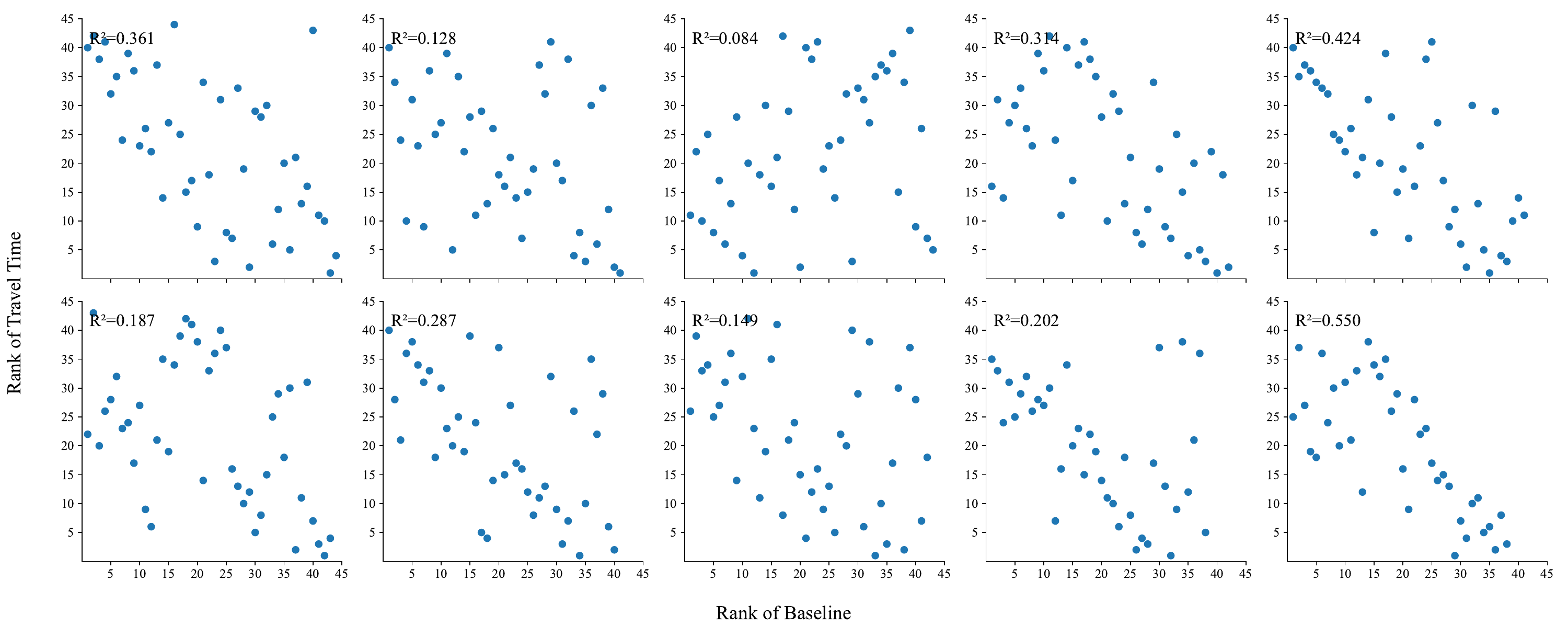}
\caption{Scatter plots of ranked time-aggregated baseline metric versus ranked travel time across all 10 calibration events. Each point represents one participant vehicle. The low $R^2$ values indicate weak alignment between the baseline metric and travel time in evaluating travel-level driving efficiency.}
\label{fig:baseline_scatter}
\end{figure}

To further compare the overall performance of PASS and the baseline metric, the rank correlation between time-aggregated values of baseline metric and travel time was computed across all 10 events (Fig. \ref{fig:baseline_scatter}). The average coefficient of determination across all events for the baseline metric is 0.269, which is significantly lower than the average $R^2$ of 0.913 for $\overline{\PASS}$. And the baseline metric exhibits negative rank correlation with travel time, further confirming its poor performance in evaluating travel-level driving efficiency. This poor performance can be attributed to its lack of consideration for the potential for speed improvement in the current lane, which is a critical aspect of driving efficiency that PASS captures through the projected attainable speed. As a result, the time-aggregated baseline metric does not align well with observed travel times. This comparison highlights the importance of incorporating a comprehensive understanding of the traffic context and potential for speed improvement in the design of instantaneous driving efficiency metrics, as exemplified by the PASS model.

\subsection{Limitations}
\label{ssec:discussion_limitations}
\subsubsection{Requirement of Lane Information}
\label{sssec:require_lane_info}
The current PASS formulation is intended for scenarios with explicit lane structure, where lane-level leading relationships can be identified. This includes standard freeway segments and lane-based structured conflict areas where virtual lanes can be defined, such as the interior of intersections when trajectories are mapped to lane-like movement channels.

In terms of lane count, the model can be applied to one-lane, two-lane, and three-lane settings by defining candidate lanes in Eq.~\eqref{eq:vproj_multi}. In this study, the practical scope is capped at a maximum of three candidate lanes around the ego vehicle and does not consider continuous multi-step lane changing within one decision horizon. In other words, PASS currently evaluates lane choices within a local lane set and is most suitable for single-step lane selection and car-following efficiency assessment under lane-based traffic organization.

\subsubsection{Fixed-Boundary Aggregation Assumption}
\label{sssec:fixed_boundary_aggregation}
PASS currently assumes a complete driving task with a clearly defined origin and destination, so that time aggregation over a full mission is well defined. Real-world operation often requires local or rolling evaluation in partial road sections without a fixed task boundary.

\subsubsection{Lack of Subjective Preference Modeling}
\label{sssec:subjective_preference}
The current PASS is an objective efficiency metric centered on rational motion performance under traffic constraints. It does not yet model subjective human preferences, which can vary widely across drivers and contexts. Specifically, aggressive and conservative drivers may have different perceptions of what constitutes efficient driving, even under identical traffic conditions.

\section{Conclusion}
Achieving high driving efficiency is a fundamental objective for autonomous vehicles (AVs), which requires a robust driving efficiency evaluation framework that can provide real-time feedback to inform decision-making. However, existing instantaneous driving efficiency metrics either neglect the influence of traffic context or lack a unified framework and clear physical interpretation. To address these limitations, this work proposes the Projected Attainable Speed Space (PASS) model, which enables a unified and physically grounded evaluation of instantaneous driving efficiency by integrating both kinematic and spatial information from the surrounding traffic environment. A driving simulation experiment was conducted, and the results demonstrate the model's ability to provide consistent driving efficiency evaluations across instantaneous and travel-level temporal scales, as evidenced by the strong alignment between time-aggregated PASS and observed travel times. The average coefficient of determination across all calibration events is 0.913, confirming the model's validity as a meaningful metric for evaluating driving efficiency in both instantaneous and travel-level contexts.

The primary contributions of this work include three key aspects: (1) the development of a conceptual framework for instantaneous driving efficiency evaluation that characterizes driving efficiency through the potential for speed improvement, quantified as the available acceleration space, and its utilization, providing a comprehensive and dynamic understanding of instantaneous driving efficiency; (2) the formulation of the PASS model, which integrates both kinematic and spatial information from the surrounding traffic environment to provide a unified and physically grounded evaluation of instantaneous driving efficiency; and (3) the demonstration of strong consistency between the time-aggregated PASS and travel-level driving efficiency metrics, confirming that the proposed model provides consistent driving efficiency evaluations across instantaneous and travel-level temporal scales.


\section{CRediT authorship contribution statement}
Xiaohua Zhao: Writing - review \& editing, Supervision, Project administration, Funding acquisition, Conceptualization. Zhaowei Huang: Writing - original draft, Methodology, Investigation, Formal analysis. Chen Chen: Writing - review \& editing, Conceptualization, Methodology, Validation, Rescource. Haiyi Yang: Visualization, Data curation.

\section{Declaration of Competing Interest}
The authors declare that they have no known competing financial interests or personal relationships that could have appeared to influence the work reported in this paper.

\section{Acknowledgements}
This paper was supported by the National Natural Science Foundation of China (No. 52572366), Beijing-Tianjin-Hebei Natural Science Foundation Cooperation Project (F2024202106), and the Young Scientists Fund of the National Natural Science Foundation of China (No. 52102411).

\section{Data availability}
The data used in this study are confidential.
 \bibliographystyle{elsarticle-num-names}
 \bibliography{citations}

@article{ahangar2021survey,
  title={A survey of autonomous vehicles: Enabling communication technologies and challenges},
  author={Ahangar, M Nadeem and Ahmed, Qasim Z and Khan, Fahd A and Hafeez, Maryam},
  journal={Sensors},
  volume={21},
  number={3},
  pages={706},
  year={2021},
  publisher={MDPI}
}

@article{ahmed2022technology,
  title={Technology developments and impacts of connected and autonomous vehicles: An overview},
  author={Ahmed, Hafiz Usman and Huang, Ying and Lu, Pan and Bridgelall, Raj},
  journal={Smart Cities},
  volume={5},
  number={1},
  pages={382--404},
  year={2022},
  publisher={MDPI}
}

@article{hang2020human,
  title={Human-like decision making for autonomous driving: A noncooperative game theoretic approach},
  author={Hang, Peng and Lv, Chen and Xing, Yang and Huang, Chao and Hu, Zhongxu},
  journal={IEEE Transactions on Intelligent Transportation Systems},
  volume={22},
  number={4},
  pages={2076--2087},
  year={2020},
  publisher={IEEE}
}

@article{zhao2024human,
  title={Human-like decision making for autonomous driving with social skills},
  author={Zhao, Chenyang and Chu, Duanfeng and Deng, Zejian and Lu, Liping},
  journal={IEEE Transactions on Intelligent Transportation Systems},
  volume={25},
  number={9},
  pages={12269--12284},
  year={2024},
  publisher={IEEE}
}

@article{karbasi2022investigating,
  title={Investigating the impact of connected and automated vehicles on signalized and unsignalized intersections safety in mixed traffic},
  author={Karbasi, Amirhosein and O'Hern, Steve},
  journal={Future transportation},
  volume={2},
  number={1},
  pages={24--40},
  year={2022},
  publisher={MDPI}
}

@article{sourelli2024modelling,
  title={Modelling the impact of context in real-world highway pull-out dynamics to inform acceptable path planning of automated vehicles},
  author={Sourelli, Anna-Maria and Welsh, Ruth and Thomas, Pete},
  journal={Transportmetrica A: transport science},
  volume={20},
  number={1},
  pages={2043951},
  year={2024},
  publisher={Taylor \& Francis}
}

@article{wang2024survey,
  title={A survey on an emerging safety challenge for autonomous vehicles: Safety of the intended functionality},
  author={Wang, Hong and Shao, Wenbo and Sun, Chen and Yang, Kai and Cao, Dongpu and Li, Jun},
  journal={Engineering},
  volume={33},
  pages={17--34},
  year={2024},
  publisher={Elsevier}
}

@article{garg2023can,
  title={Can connected autonomous vehicles improve mixed traffic safety without compromising efficiency in realistic scenarios?},
  author={Garg, Mohit and Bouroche, M{\'e}lanie},
  journal={IEEE Transactions on Intelligent Transportation Systems},
  volume={24},
  number={6},
  pages={6674--6689},
  year={2023},
  publisher={IEEE}
}

@article{zhao2018platoon,
  title={A platoon based cooperative eco-driving model for mixed automated and human-driven vehicles at a signalised intersection},
  author={Zhao, Weiming and Ngoduy, Dong and Shepherd, Simon and Liu, Ronghui and Papageorgiou, Markos},
  journal={Transportation Research Part C: Emerging Technologies},
  volume={95},
  pages={802--821},
  year={2018},
  publisher={Elsevier}
}

@article{guo2019joint,
  title={Joint optimization of vehicle trajectories and intersection controllers with connected automated vehicles: Combined dynamic programming and shooting heuristic approach},
  author={Guo, Yi and Ma, Jiaqi and Xiong, Chenfeng and Li, Xiaopeng and Zhou, Fang and Hao, Wei},
  journal={Transportation research part C: emerging technologies},
  volume={98},
  pages={54--72},
  year={2019},
  publisher={Elsevier}
}

@article{han2020energy,
  title={Energy-aware trajectory optimization of CAV platoons through a signalized intersection},
  author={Han, Xiao and Ma, Rui and Zhang, H Michael},
  journal={Transportation Research Part C Emerging Technologies},
  volume={118},
  pages={102652},
  year={2020},
  publisher={Elsevier}
}

@article{chen2020optimal,
  title={Optimal control for connected and autonomous vehicles at signal-free intersections},
  author={Chen, Boli and Pan, Xiao and Evangelou, Simos A and Timotheou, Stelios},
  journal={IFAC-PapersOnLine},
  volume={53},
  number={2},
  pages={15306--15311},
  year={2020},
  publisher={Elsevier}
}

@article{pourmehrab2020signalized,
  title={Signalized intersection performance with automated and conventional vehicles: A comparative study},
  author={Pourmehrab, Mahmoud and Emami, Patrick and Martin-Gasulla, Marilo and Wilson, Jabari and Elefteriadou, Lily and Ranka, Sanjay},
  journal={Journal of Transportation Engineering, Part A: Systems},
  volume={146},
  number={9},
  pages={04020089},
  year={2020},
  publisher={American Society of Civil Engineers}
}

@article{li2022analysis,
  title={An analysis of the value of optimal routing and signal timing control strategy with connected autonomous vehicles},
  author={Li, Tang and Guo, Fangce and Krishnan, Rajesh and Sivakumar, Aruna},
  journal={Journal of Intelligent Transportation Systems},
  volume={28},
  number={2},
  pages={252--266},
  year={2022},
  publisher={Taylor \& Francis}
}

@article{mohebifard2021connected,
  title={Connected automated vehicle control in single lane roundabouts},
  author={Mohebifard, Rasool and Hajbabaie, Ali},
  journal={Transportation research part C: emerging technologies},
  volume={131},
  pages={103308},
  year={2021},
  publisher={Elsevier}
}

@article{mohebifard2022trajectory,
  title={Trajectory control in roundabouts with a mixed fleet of automated and human-driven vehicles},
  author={Mohebifard, Rasool and Hajbabaie, Ali},
  journal={Computer-Aided Civil and Infrastructure Engineering},
  volume={37},
  number={15},
  pages={1959--1977},
  year={2022},
  publisher={Wiley Online Library}
}

@article{malikopoulos2018optimal,
  title={Optimal control for speed harmonization of automated vehicles},
  author={Malikopoulos, Andreas A and Hong, Seongah and Park, B Brian and Lee, Joyoung and Ryu, Seunghan},
  journal={IEEE Transactions on Intelligent Transportation Systems},
  volume={20},
  number={7},
  pages={2405--2417},
  year={2018},
  publisher={IEEE}
}

@article{dai2017effect,
  title={The effect of connected vehicle environment on global travel efficiency and its optimal penetration rate},
  author={Dai, Rongjian and Lu, Yingrong and Ding, Chuan and Lu, Guangquan},
  journal={Journal of Advanced Transportation},
  volume={2017},
  number={1},
  pages={2697678},
  year={2017},
  publisher={Wiley Online Library}
}

@article{sun2023optimal,
  title={Optimal control of connected autonomous vehicles in a mixed traffic corridor},
  author={Sun, Wenbo and Zhang, Fangni and Liu, Wei and He, Qingying},
  journal={IEEE Transactions on Intelligent Transportation Systems},
  volume={25},
  number={5},
  pages={4206--4218},
  year={2023},
  publisher={IEEE}
}

@article{oh2024exploring,
  title={Exploring Driving Parameter Settings for Autonomous Vehicles: Considering Travel Efficiency and Safety in Urban Traffic Environments},
  author={Oh, Gyungtaek and Lee, Eugene and Kim, Hyungjoo and Hu, Jia and Yun, Ilsoo and Ko, Hangeom and Cho, Sungwoo and So, Jaehyun},
  journal={IEEE Transactions on Intelligent Vehicles},
  year={2024},
  publisher={IEEE}
}

@article{xu2018reinforcement,
  title={A reinforcement learning approach to autonomous decision making of intelligent vehicles on highways},
  author={Xu, Xin and Zuo, Lei and Li, Xin and Qian, Lilin and Ren, Junkai and Sun, Zhenping},
  journal={IEEE Transactions on Systems, Man, and Cybernetics: Systems},
  volume={50},
  number={10},
  pages={3884--3897},
  year={2018},
  publisher={IEEE}
}

@article{lv2022safe,
  title={A safe and efficient lane change decision-making strategy of autonomous driving based on deep reinforcement learning},
  author={Lv, Kexuan and Pei, Xiaofei and Chen, Ci and Xu, Jie},
  journal={Mathematics},
  volume={10},
  number={9},
  pages={1551},
  year={2022},
  publisher={MDPI}
}

@article{hang2021cooperative,
  title={Cooperative decision making of connected automated vehicles at multi-lane merging zone: A coalitional game approach},
  author={Hang, Peng and Lv, Chen and Huang, Chao and Xing, Yang and Hu, Zhongxu},
  journal={IEEE Transactions on Intelligent Transportation Systems},
  volume={23},
  number={4},
  pages={3829--3841},
  year={2021},
  publisher={IEEE}
}

@article{yang2023multi,
  title={Multi-lane coordinated control strategy of connected and automated vehicles for on-ramp merging area based on cooperative game},
  author={Yang, Lan and Zhan, Jiahao and Shang, Wen-Long and Fang, Shan and Wu, Guoyuan and Zhao, Xiangmo and Deveci, Muhammet},
  journal={IEEE Transactions on Intelligent Transportation Systems},
  volume={24},
  number={11},
  pages={13448--13461},
  year={2023},
  publisher={IEEE}
}

@article{cai2024game,
  title={Game-theoretic decision-making method and motion planning for autonomous vehicles in overtaking},
  author={Cai, Lei and Guan, Hsin and Xu, Qi Hong and Jia, Xin and Zhan, Jun},
  journal={IEEE Transactions on Intelligent Transportation Systems},
  volume={25},
  number={8},
  pages={9693--9709},
  year={2024},
  publisher={IEEE}
}

@article{liu2019novel,
  title={A novel lane change decision-making model of autonomous vehicle based on support vector machine},
  author={Liu, Yonggang and Wang, Xiao and Li, Liang and Cheng, Shuo and Chen, Zheng},
  journal={IEEE access},
  volume={7},
  pages={26543--26550},
  year={2019},
  publisher={IEEE}
}

@inproceedings{shi2019driving,
  title={Driving decision and control for automated lane change behavior based on deep reinforcement learning},
  author={Shi, Tianyu and Wang, Pin and Cheng, Xuxin and Chan, Ching-Yao and Huang, Ding},
  booktitle={2019 IEEE intelligent transportation systems conference (ITSC)},
  pages={2895--2900},
  year={2019},
  organization={IEEE}
}

@article{lu2024game,
  title={Game-theoretic lane change decision-making method considering traffic trend},
  author={Lu, Xinghao and Zhao, Haiyan and Li, Cheng and Liu, Wan and Gao, Bingzhao and Zhou, Qiuzhan},
  journal={IEEE Transactions on Industrial Electronics},
  volume={71},
  number={11},
  pages={14793--14802},
  year={2024},
  publisher={IEEE}
}

@inproceedings{deng2022lane,
  title={Lane change decision-making with active interactions in dense highway traffic: A Bayesian game approach},
  author={Deng, Zejian and Hu, Wen and Yang, Yanding and Cao, Kai and Cao, Dongpu and Khajepour, Amir},
  booktitle={2022 IEEE 25th International Conference on Intelligent Transportation Systems (ITSC)},
  pages={3290--3297},
  year={2022},
  organization={IEEE}
}

@article{bian2021influence,
  title={Influence of prompt timing and messages of an audio navigation system on driver behavior on an urban expressway with five exits},
  author={Bian, Yang and Zhang, Xiaolong and Wu, Yiping and Zhao, Xiaohua and Liu, Hao and Su, Yuelong},
  journal={Accident Analysis \& Prevention},
  volume={157},
  pages={106155},
  year={2021},
  publisher={Elsevier}
}






\end{document}